\documentclass[journal]{IEEEtran}
\usepackage[switch]{lineno}
\newcommand*\patchAmsMathEnvironmentForLineno[1]{
  \expandafter\let\csname old#1\expandafter\endcsname\csname #1\endcsname
  \expandafter\let\csname oldend#1\expandafter\endcsname\csname end#1\endcsname
  \renewenvironment{#1}
     {\linenomath\csname old#1\endcsname}
     {\csname oldend#1\endcsname\endlinenomath}}
\newcommand*\patchBothAmsMathEnvironmentsForLineno[1]{
  \patchAmsMathEnvironmentForLineno{#1}
  \patchAmsMathEnvironmentForLineno{#1*}}
\AtBeginDocument{
\patchBothAmsMathEnvironmentsForLineno{equation}
\patchBothAmsMathEnvironmentsForLineno{align}
\patchBothAmsMathEnvironmentsForLineno{flalign}
\patchBothAmsMathEnvironmentsForLineno{alignat}
\patchBothAmsMathEnvironmentsForLineno{gather}
\patchBothAmsMathEnvironmentsForLineno{multline}
}
\usepackage{cite}                      % needed to automatically sort the references
\usepackage{tabu}                      % only used for the table example
\usepackage{booktabs}                  % only used for the table example
\usepackage{subfig}
\usepackage{times}
\usepackage{epsfig}
\usepackage{graphicx}
\usepackage{amsmath}
\usepackage{amssymb}
\usepackage{algorithm}
\usepackage{color}
\usepackage{bbm}
\usepackage{bm}
\usepackage{multirow}
\usepackage{comment}
\usepackage{here}
\usepackage{booktabs}
\usepackage{arydshln}
\usepackage{hyperref}
\usepackage{float}
\usepackage[noend]{algpseudocode} 
\usepackage{algorithmicx}
\usepackage[utf8]{inputenc}

\ifCLASSINFOpdf
\else
\fi
\begin{document}
\title{Frequency-Guided Multi-Level Human Action Anomaly Detection with Normalizing Flows}

\author{Shun~Maeda*,
        Chunzhi~Gu*,
        Jun~Yu,
        Shogo~Tokai,
        Shangce~Gao,
        and~Chao~Zhang% <-this % stops a space
% \thanks{C. Gu* and S. Kuriyama are with the Department of Computer Science and Engineering, Toyohashi University of Technology, Toyohashi, Japan (e-mails: gu@cs.tut.ac.jp, sk@tut.jp).}% <-this % stops a space
% \thanks{C. Zhang is with the School of Engineering, University of Fukui, Fukui, Japan (zhang@u-fukui.ac.jp).}
\thanks{S. Maeda* and S. Tokai are with the School of Engineering, University of Fukui, Fukui, Japan (msd24006@u-fukui.ac.jp, tokai@u-fukui.ac.jp).}% <-this % stops a space
\thanks{C. Gu* is with the Department of Computer Science and Engineering, Toyohashi University of Technology, Toyohashi, Japan (gu@cs.tut.ac.jp).}% <-this % stops a space
\thanks{J. Yu is with Institute of Science and Technology, Niigata University, Niigata, Japan (yujun@ie.niigata-u.ac.jp).}% <-this % stops a space
\thanks{S. Gao and C. Zhang are with the Faculty of Engineering, University of Toyama, Toyama, Japan (gaosc@eng.u-toyama.ac.jp, zhang@eng.u-toyama.ac.jp).}% <-this % stops a space
\thanks{*Equal contribution.}% <-this % stops a space
}
\maketitle

\begin{abstract}
We introduce the task of human action anomaly detection (HAAD), which aims to identify anomalous motions in an unsupervised manner given only the pre-determined normal category of training action samples. Compared to prior human-related anomaly detection tasks which primarily focus on unusual events from videos, HAAD involves the learning of specific action labels to recognize semantically anomalous human behaviors. To address this task, we propose a normalizing flow (NF)-based detection framework where the sample likelihood is effectively leveraged to indicate anomalies. As action anomalies often occur in some specific body parts, in addition to the full-body action feature learning, we incorporate extra encoding streams into our framework for a finer modeling of body subsets. Our framework is thus multi-level to jointly discover global and local motion anomalies. Furthermore, to show awareness of the potentially jittery data during recording, we resort to discrete cosine transformation by converting the action samples from the temporal to the frequency domain to mitigate the issue of data instability. Extensive experimental results on two human action datasets demonstrate that our method outperforms the baselines formed by adapting state-of-the-art human activity AD approaches to our task of HAAD. 
\end{abstract}


% \fi
\IEEEpeerreviewmaketitle

\section{Introduction}

\IEEEPARstart{A}{nomaly} detection (AD) enables the recognition of whether an input sample meets the expected industrial requirements or not. It plays a key role in various fields such as biomedical analysis \cite{yoo2018deep} or manufacturing detection \cite{wang2021student}. In recent years, the field of AD has been developed primarily on 2D images \cite{schlegl2017unsupervised,ristea2022self,lu2018anomaly} to fulfill the broad needs of real-world applications. 

In addition to images, AD has also progressed in the direction of human activity, which aims to discover unusual events from the recorded video data \cite{park2020learning,doshi2020continual,ristea2022self}. Generally, similar to image AD, video AD also follows the unsupervised learning diagram by extracting anomaly-free motion features from the observed normal activities. Earlier attempts \cite{li2013anomaly,jiang2011anomalous} directly perform AD on the raw video data. As pointed out in \cite{sabokrou2017deep}, raw video data can include irrelevant contexts, such as background or illumination variations, that impose a negative influence on the detection accuracy. Therefore, though sparsely studied, some recent approaches \cite{morais2019learning,rodrigues2020multi, flaborea2023multimodal} have shifted towards using skeletal representations to focus straightforwardly on human motion itself. Flaborea et al. \cite{flaborea2023multimodal} measured the reconstruction error with the diffusion generative model to identify the anomaly. However, reconstruction-based methods with such a powerful generative model tend to over-generalize that even anomalous samples can be decently recovered.
For example, Hirschorn et al. \cite{hirschorn2023normalizing} proposed regarding the likelihood derived with the normalizing flow model as the anomaly score for detection. Despite the effectiveness, since it is developed to capture the general 2D human event abnormality, the likelihood can be sensitive to the movements with similar global motion trends. 

%introduces a normalizing flow approach to work with space-time graph data and score the likelihood instead of relying on reconstruction error. Although they also use normalizing flow as in our study, they might misidentify resemble movements (e.g., run and walk) since they do not take into account the fine movements of skeleton parts.

\begin{figure}[t]     
    \centering     
    \includegraphics[scale=0.28]{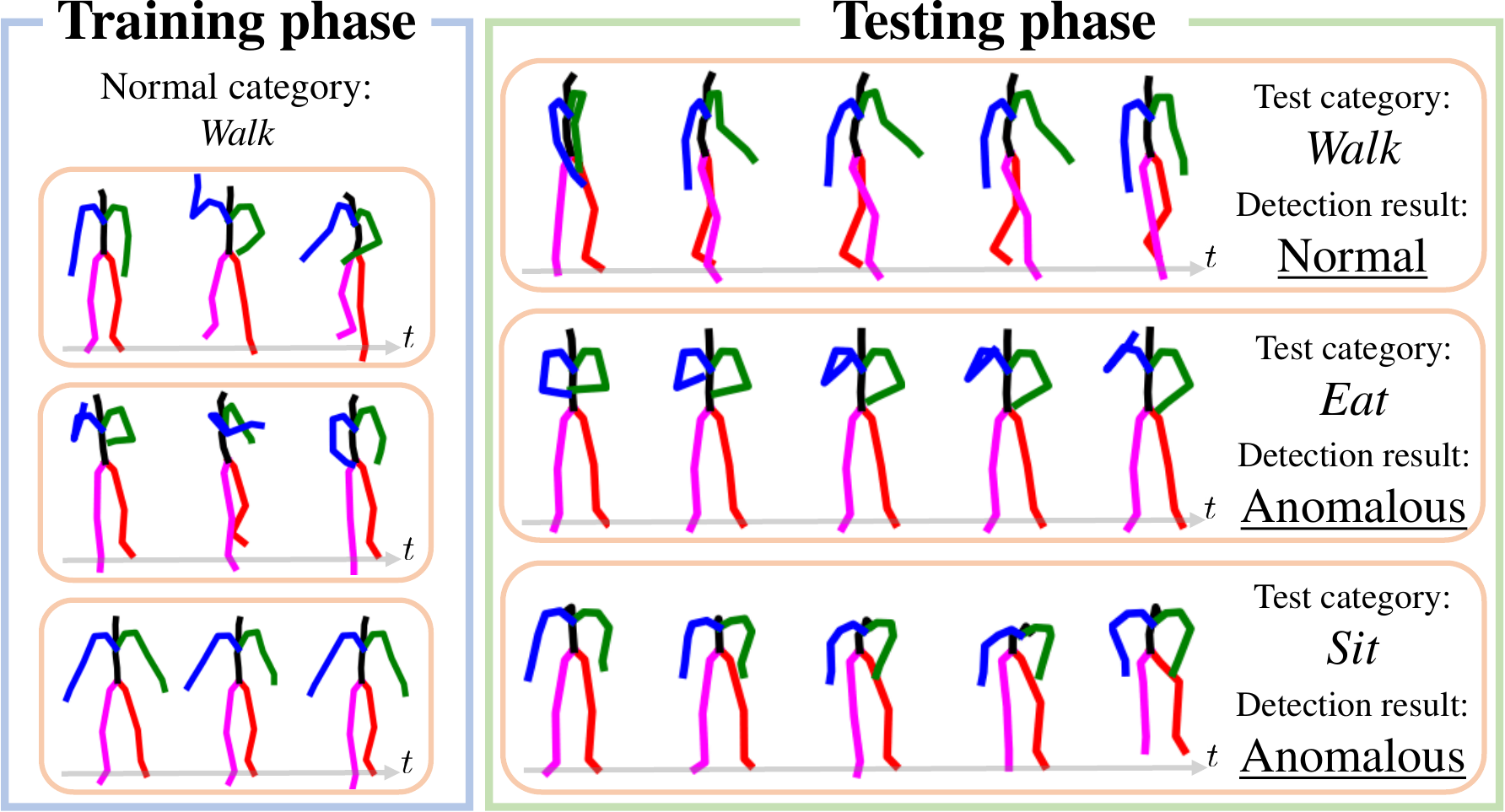}     
    \caption{\textbf{Human action anomaly detection.} Given only the selected normal action class (blue) for training, our method detects whether an arbitrary test motion sample is anomalous or not (green).     Each orange box visualizes a human motion in the temporal domain from left to right.}     \label{fig:top-page}
\end{figure}

Further, prior human-oriented approaches do not show awareness of the semantics of human actions for AD. It should be noted that identifying anomalies in the semantic level is crucial for many safety-practical applications. For example, in real-world construction sites or factories, it is important to ensure that the workers should only remain \textit{walking} in some certain areas, rather than taking any other actions, such as \textit{running} or \textit{sitting}, to avoid potential dangers. Compared to conventional semantic-free AD, introducing specific action labels for AD requires to model the underlying semantic features of the given action type. This also involves accurately distinguishing spatial-temporally analogous actions with different semantic labels, which is more challenging and remains mostly unexplored. As an alternative solution, human action recognition \cite{shuchang2022survey, pang2023skeleton, yang2022recurring} also aims at classifying the given action into several pre-determined motion categories. However, because constructing a dataset that covers the diverse space of possible actions in real-world industrial scenarios is virtually unreasonable, the training of action recognition model can be highly complicated.

Considering the challenges described above, in this paper, we propose a new task, human action anomaly detection (HAAD), which aims to detect anomalous 3D motion patterns with a target normal category of human action in an unsupervised manner, as displayed in Fig. \ref{fig:top-page}. It differs from previous human activity AD tasks in that we introduce the exact semantic action type to indicate the normal or anomalous data category. Importantly, despite the same action label, samples in the normal set can express huge diversity to account for the stochastic nature of human motion, which imposes challenges during detection. To this end, we propose a novel normalizing flow (NF)-based framework to learn quality motion representations for HAAD. Since the recorded 3D human motion can include noise during recording, the motion data  can suffer from different degrees of jitter that hinders  detection accuracy. To address this issue, instead of directly learning on the temporal domain as in the prior approach \cite{luo2021normal}, we propose handling HAAD in the frequency domain by performing Discrete Cosine Transform (DCT) on the input sequential motion sample. In particular, inspired by \cite{mao2019learning}, we only adopt the low-frequency DCT components to ensure temporal trajectory smoothness. This allows us to remove the instability within the motion data that complicates the detection.

Moreover, we notice that human motion can only involve local differences in some key body parts to perform different actions. For example, one would only enforce upper-body movements to perform some locally static actions, like \textit{phoning} or \textit{drinking}, while keeping his/her lower body still. To show awareness of this human motion attribute, in addition to learning the full-body motion, we formulate a multi-level stream by separating the human body into several subsets such that our model can also characterize the anomalies within some specific body parts. We further utilize the graph convolutional network (GCN) to capture the spatial dependencies of the skeletal presentations for human pose sequences. Consequently, our framework enables extracting local and global motion features jointly to detect subtle action anomalies.

Given the multi-level motion features obtained from the target action class specified as normal via GCN, the NF then learns to maximize the likelihood of the samples within this action category such that the reversible mappings can realize generation. In the inference phase, we draw inspiration from previous image AD schemes \cite{reiss2021panda} by performing $K$-nearest neighbor (KNN) search on the feature vector regressed via the NF to endow each test motion sample with an anomaly score. Since the test samples with anomalous action categories are unseen during optimization, it would induce a large drop in the anomaly score to indicate abnormality, while the normal test samples would be scored high. As will be shown in our experiments, our KNN-based action anomaly scoring contributes to a higher detection accuracy and robustness compared to the one \cite{hirschorn2023normalizing} that straightforwardly exploits the likelihood as anomaly scores.

Our contributions can be thus summarized as follows: (i) We introduce a new task, human action anomaly detection, which regards the anomaly as specific action categories for human motion; (ii) We propose to address this task under a novel frequency-guided detection framework formulated by normalizing flow; (iii) We incorporate a multi-level detection pipeline into our model to facilitate a better learning of local anomalous action patterns.

Extensive experimental results and ablative evaluations on two large-scale human motion datasets demonstrate that our method outperforms other baseline approaches constructed by extending state-of-the-art human event AD models to our task.

\section{Related Work}
\begin{figure*}[t]
    \centering
    \includegraphics[width=0.8\linewidth]{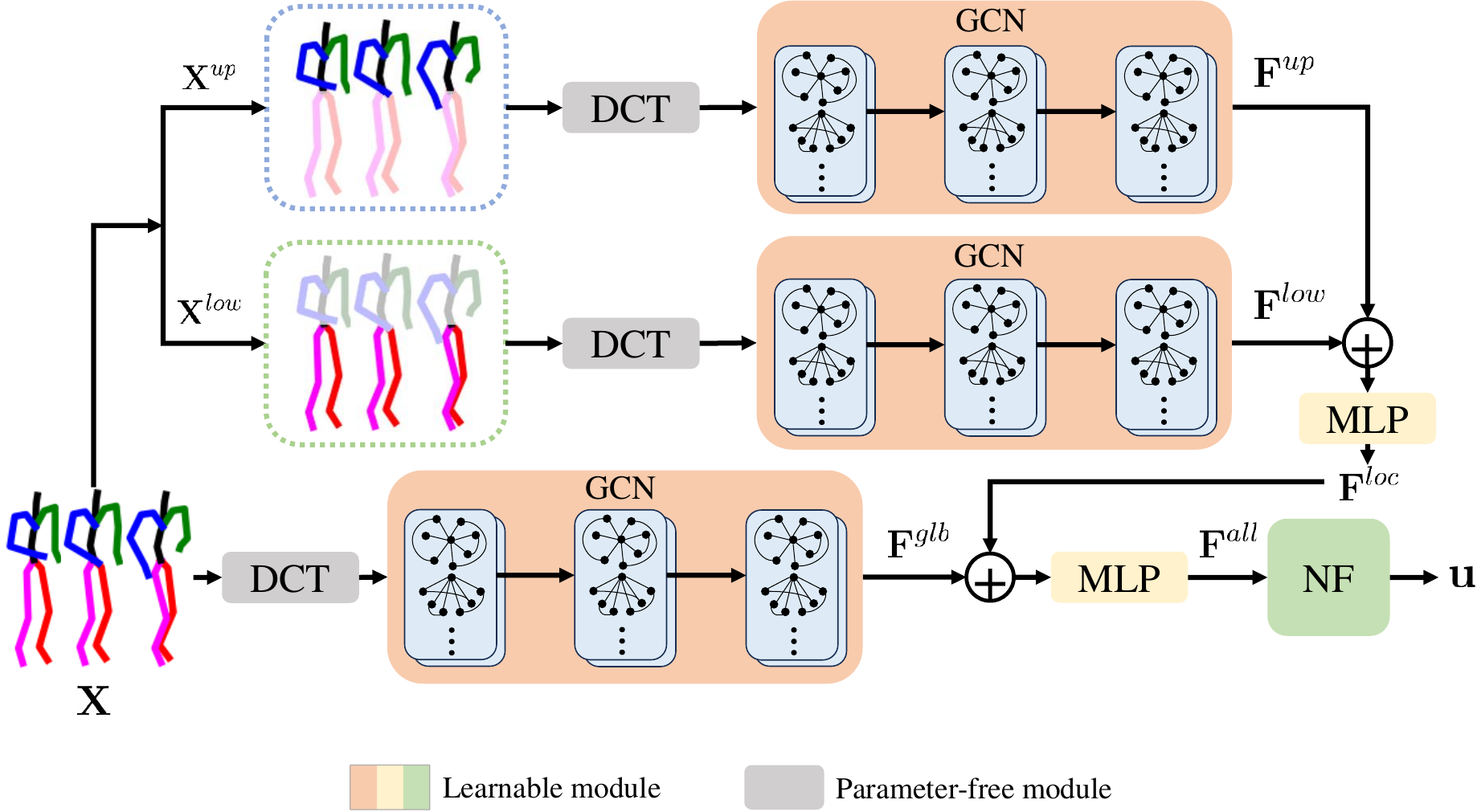}
    \caption{\textbf{Overview} of our multi-level NF-based human action anomaly learning framework.}
    \label{fig:Our_approach}
\end{figure*}

In this section, we first review the previous human-related AD techniques. We then discuss some action recognition methods. Eventually, we review prior AD techniques for time series data.

\subsection{Human-Related Anomaly Detection}
The field of human-related AD mostly aims to detect unexpected events from the recorded videos, which is referred to as video AD \cite{9753580,georgescu2021anomaly,morais2019learning,georgescu2021background,li2013anomaly,wang2018video}. %Early attempts \cite{jiang2011anomalous,tung2011goal} relied on hand-craft feature extraction approaches, such as Gaussian Mixture Models \cite{tung2011goal} or Hidden Markov Model \cite{jiang2011anomalous}. In general, these methods show poor robustness to unknown or unseen types of anomalies \cite{hu2016video}. 
Recent efforts generally resort to the powerful latent representations using deep neural networks. Typically, the powerful deep representations are usually derived via proxy tasks, such as self-reconstruction \cite{georgescu2021background,ristea2022self,wang2022video}. Ristea et al. \cite{ristea2022self} introduced a self-supervised framework with the dilated convolution and channel attention mechanism to discover anomalous patterns.  Wang et al. \cite{wang2022video}  decoupled the video AD task into the spatial and temporal jigsaw puzzles to model normal appearance and motion patterns, respectively. 
However, because the anomalies are identified based on the image sequences, the change of illuminations or scene contexts in the background can induce erroneous detection results. As such, more recently, human AD has been advanced towards using skeletal representations to circumvent undesired factors. In this regard, the most closely related works to ours are \cite{hirschorn2023normalizing,flaborea2023multimodal}, which involve skeleton-based AD. Specifically, Hirschorn et al. \cite{hirschorn2023normalizing} proposed a normalizing flows-based framework for detecting anomalies in human pose data by learning the distribution of normal poses and identifying low-likelihood poses as anomalies. Flaborea et al. \cite{flaborea2023multimodal} proposed a multimodal diffusion-based model that first learns the distribution of normal skeleton motion, and then identifies anomalies by measuring the reconstruction error between the generated and the input skeletal sequence. In principle, similar to video AD for humans, these two methods aim to recognize the general motion anomalies without considering the specific anomalous actions. As such, they lack a finer semantic-level modeling for human actions, leading to low capacity in identifying subtle anomalies within subtle actions.

\subsection{Human Action Recognition}
Human action recognition \cite{shuchang2022survey} remains to date as the main technique to identify the category of the given motion clip. These methods \cite{yang2022recurring,hara2018can,feichtenhofer2016convolutional,bruce2022mmnet,duan2022revisiting,caetano2019skelemotion} usually temporally and spatially extract discriminative representations from the skeleton sequences to classify the given input action. Typically, their encoding backbones are selected as architectures with a strong capacity to handle human skeletal data. Lee et al. \cite{lee2017ensemble} proposed a Temporal Sliding LSTM (TS-LSTM) where multi-term LSTMs are leveraged to yield robustness to variable temporal dynamics. Zhang et al. \cite{zhang2019view} devised a viewpoint adaptation strategy via CNNs to determine the most suitable observation viewpoint such that temporal features can be best exploited to recognize the actions. Despite the effectiveness, these methods mainly explore the temporal relationship and tend to overlook the spatial dependencies of human actions. Later approaches \cite{yan2018spatial, li2019actional, pang2023skeleton,kim2023cross} then attempt to harmonize the spatial and temporal features to provide better solutions. Yan et al. \cite{yan2018spatial} constructed a series of spatial-temporal graph convolution layers to model the spatial configuration within the dynamic motion information. To discover richer dependencies among joints, Li et al. \cite{li2019actional} further proposed action-structural graph convolution blocks for higher order relationship modeling. Pang et al. \cite{pang2023skeleton} designed a contrastive network that combines a two-stream spatial-temporal network to capture the relationship between arbitrary joints within and across frames. In general, human action recognition shares the closest motivation to our task in understanding 3D human action semantics. However, it requires the collection of annotated motion data with diverse action categories to provide the required supervision. Differently, our task only demands the selected one type of action for one-class classification, which is less labor-intensive and readily applicable to real-world tasks. %Shah et al. \cite{shah2023multi} introduced a multi-view framework that leverages contrastive learning and cross-view consistency for achieving robustness to viewpoints.} 

\subsection{Anomaly Detection for Time series Data}
Besides the mainstream AD methods \cite{roth2022towards,defard2021padim,cohen2020sub} devised for 2D images to meet the industrial demands for manufacturing, AD is also broadly studied for time-series data. Specifically, time series AD aims to discover 
unusual data behaviors at some specific time steps. Xu et al. \cite{xu2022anomaly} devised an anomaly-attention mechanism to compute the association discrepancy by amplifying the normal-abnormal distinguishability with association-based detection criterion.
Wong et al. \cite{wong2022aer} fused the auto-encoding and LSTM to jointly harvest the strength of prediction-based and reconstruction-based models for time series anomalies. Zhou et al. \cite{zhou2023detecting} modeled the interdependencies into the dynamic graph to capture the complex dependencies and correlations within multivariate time series data.  Although our task also involves detecting anomalies within time-series human motion data, we aim to discover the global action anomalies from complex human motion patterns, instead of localizing the anomalous time-steps. Regarding this, HAAD differs fundamentally from prior time-series AD tasks.

\section{Method}
Let us now introduce our approach to one-class human action anomaly detection. Formally, we define a set of motion data as $\mathcal{X}^{c} = \{ \mathbf{X}^{c}_1, \cdots, \mathbf{X}^{c}_m, \cdots, \mathbf{X}^{c}_{N_c}\}$ for the $c$-th action category with $N_c$ motion clips, where each sample $\mathbf{X}^{c}_m =  [\mathbf{x}^c_1, \cdots, \mathbf{x}^c_h, \cdots, \mathbf{x}^c_H]^T \in \mathbb{R}^{P \times H}$ is composed of $H$ frames of $P$-dimensional human poses $\mathbf{x}^c_h\in \mathbb{R}^{P}$. Given $\mathcal{X}^{c}$ as the only available training data, our goal is to identify whether the action category ${c_u}$ of an arbitrary unseen test motion clip $\mathbf{Y}^{c_u} \in \mathbb{R}^{P \times H}$ matches the desired action type ${c}$. As illustrated in Fig. \ref{fig:Our_approach}, our method involves a multi-level architecture to learn semantic action features, which are eventually fed into the normalizing flow model to optimize the sample likelihood for action AD. We detail each component of our method in the following of this section.

%Let us introduce our approach to HCAD. An overview of the proposed method is shown in Fig. \ref{fig:Our_approach}. Our goal is to identify whether the motion is normal or not normal. Our approach involves representing a human pose as a 3D joint coordinate in a frame, denoted by the vector $\mathbf{x}_i\in \mathbb{R}^{K} $, with $K$ as the joint node in the $i$-th frame and a human motion sequence $ \mathbf{X} = [\mathbf{x}_1, \mathbf{x}_2, ..., \mathbf{x}_H]^T $, where $H$ is frame length.
% Kは関節数ではなく，3 x 関節数を表したい

\subsection{Motion Feature Learning}
\noindent\textbf{Frequency-guided encoding.}
We aim at learning quality deep features to detect semantic anomalies in human actions. One straightforward way is to use temporal encoding modules (e.g., Recurrent Neural Networks) to embed the flow of the sequential input. However, since the recorded human motion data can contain different degrees of noise, including jittering or flipping, directly learning on the temporal domain can mislead the detector into regarding such instability as an anomaly. To pursue a more suitable representation, we resort to frequency guidance by applying the DCT transform on each sample in $\mathcal{X}^{c}$ for temporal encoding. Specifically, for an arbitrary human motion $\mathbf{X} \in \mathbb{R}^{P \times H}$, the DCT transform in performed via
\begin{equation}
\label{Formula: DCT transform}
\mathbf{C} = \mathbf{X} \mathbf{T},
\end{equation}
where  $ \mathbf{T} \in \mathbb{R}^{H \times M} $ denotes a predefined DCT basis and $ \mathbf{C} \in \mathbb{R}^{P \times M} $ refers to the first $M$ DCT coefficients. Each row of $ \mathbf{C}$ constitutes the DCT coefficients of one joint coordinate/rotation sequence. Importantly, by discarding some high-frequency DCT basis, the original motion can be compactly represented in a smoother manner. Hence, we set $M$ small to only retain low frequencies to mitigate the negative influence on detection caused by jittery motion, and use the extracted $\mathbf{C}$ to facilitate further deep motion learning. We next need to discuss how capture the joint dependencies for  spatial embedding.

\noindent \textbf{Graph Convolution.}
Given the compact frequency representation  $\mathbf{C}$ expressed by DCT coefficients, we draw inspiration from \cite{pang2023skeleton, mao2019learning} by leveraging Graph Convolutional Networks (GCNs) to characterize spatial dependencies among the human joints. Let the representation of the human body be a fully-connected graph comprising $P$ nodes. By defining a GCN with a total of $L$ layers, we consider that the input to the $l\in(1,L)$-th graph convolution layer is a matrix $ \mathbf{F}^{(l)} \in \mathbb{R} ^ {P \times D} $, where $D$ is the output feature dimension of the previous layer. For the $(l+1)$-th layer, the graph convolution computes the feature $ \mathbf{F}^{(l+1)} \in \mathbb{R}^{P \times \hat{D}} $ as 
\begin{equation}
\label{Formula: GCN layer l}
\mathbf{F}^{(l+1)} = \sigma(\mathbf{A}^{(l)}\mathbf{F}^{(l)}\mathbf{W}^{(l)}), \end{equation}
where $ \mathbf{A}^{(l)} \in \mathbb{R}^{P \times P} $ is a weighted adjacency matrix that represents the connection strength of the edges in the graph and $ \mathbf{W}^{(l)} \in \mathbb{R}^{D \times \hat{D}} $ denotes the matrix of trainable weights with $\hat{D}$ being the layer feature dimension. Following \cite{kipf2016semi}, instead of using a pre-determined connectivity, we make $\mathbf{A}^{(l)}$ learnable to capture the dependencies among different joint trajectories. The regressed graph feature is further activated by $ \sigma (\cdot) $ to derive the  $(l+1)$-th layer output. The first layer directly takes the $P \times M$ DCT coefficients matrix $\mathbf{C}$ as input, and eventually, our GCN produces $\mathbf{F}^{(L)}$ as the output which encodes the spatial structure of human poses.

\noindent \textbf{Multi-level action feature learning.}
We note that although our GCN-based  spatial encoding is effective in modeling the global human behavior, the anomaly in human action can often occur in fine local body parts. More specifically, even if two semantic action labels are perceived and understood differently, these two motions can overlap noticeably in the global level. For example, for a standing person performing the actions of \textit{waving} and \textit{drinking}, the difference can focus sorely on the upper body, with the lower body being still. To reflect this intuition into our approach, in addition to the whole body learning stream for encoding global features, we propose to introduce two subset GCN streams to characterize local body movements. In particular, we divide the fully body motion into two folds: $\mathbf{X} = \{\mathbf{X}^{up}, \mathbf{X}^{low}\}$, which represent the upper $\mathbf{X}^{up} \in \mathbb{R}^{P^{up} \times H}$ and lower $\mathbf{X}^{low} \in \mathbb{R}^{P^{low} \times H}$
body motions with $P^{up}$ and $P^{low}$ dimensions, respectively. Similar to the full body scenario in Eq. \ref{Formula: DCT transform}, we first apply DCT to $\mathbf{X}^{up}$ and $\mathbf{X}^{low}$, and then prepare two GCNs to learn from the corresponding frequency encodings. The two subset GCN streams output $\mathbf{F}^{up}$ and $\mathbf{F}^{low}$ as the subset motion embeddings, respectively. The embeddings are then concatenated to pass through a multi-layer perceptron (MLP) for feature fusing: 
$\mathbf{F}^{loc} = \text{MLP}(\mathbf{F}^{up} \oplus \mathbf{F}^{low})$.

Importantly, because the subset GCNs only have access to one body portion, $\mathbf{F}^{loc}$ serves as a strong semantic guidance to learn the local action attributes. By denoting the feature for the full-body human action obtained in the global branch as $\mathbf{F}^{glb}$, we derive the final fused feature in the graph convolution stage with another MLP: $\mathbf{F}^{all} = \text{MLP}(\mathbf{F}^{glb} \oplus \mathbf{F}^{loc})$. 
Our pipeline is therefore multi-level in its design that facilitates exploring both local and global action modes. As will be shown in our experiments in Sec. \ref{Table. HumanAct12 multi-level}\ref{Table. UESTC multi-level}, the multi-level architecture allows our framework to detect subtle action anomalies with similar semantic behaviors. 
We rewrite $\mathbf{F}^{all}$ to $\mathbf{F}$ for brevity in the following of this section. We next need to know how to exploit $\mathbf{F}$ for action AD.

\subsection{Normalizing Flow for action anomaly detection}
Since our goal is to identify the action anomaly within human motion, we need to compute the anomaly score to facilitate the detection. To this end, given the extracted quality motion embedding $\mathbf{F}$, we propose to model the resulting abnormality via normalizing flow (NF).

\noindent \textbf{Training.} NF is a type of generative model that maps a sample in the motion feature distribution $ p(\mathbf{F})$ to a latent representation $ \mathbf{u} = f(\mathbf{F}) $ which follows a simple Gaussian distribution $ \mathbf{u}\sim\mathcal{N}(0, \mathbf{I}) $. The mapping $f$ is formulated with a series of deep neural networks to ensure expressive generation capability. The training of NF aims to optimize the likelihood of each sample, following: 
\begin{equation}
\label{Formula: Normalizing Flow}
p( \mathbf{F}) = q( \mathbf{u}) \left| \operatorname{det} \left( \frac{ \partial f}{ \partial \mathbf{F}} \right) \right|,
\end{equation}
where $ q( \mathbf{u}) = \mathcal{N}(\mathbf{u}|0, \mathbf{I}) $ and $ \operatorname{det} \left( \frac{ \partial f}{ \partial \mathbf{F}} \right) $ denotes the  determinant of the Jacobian matrix of $ f(\cdot) $.

Our key motivation of using NF is that, different from other forms of generative models, such as VAEs \cite{kingma2013auto} or GANs \cite{goodfellow2014generative},  
the output of NF (i.e., $f$) measures the exact likelihood of each sample. Therefore, we can train the NF model on a pre-determined class of normal actions, and in the testing phase, the motion samples with any other action categories would induce a huge likelihood drop to indicate anomaly. 

Formally, given the feature vector $\mathbf{F}^c$ obtained on the sample in the normal action category $c$, our NF model learns to \textit{minimize} the negative log-likelihood (NLL) as the training objective:
\begin{align}
\label{Formula: loss log-likelihood}
\mathcal{L} & = -\log p(\mathbf{F}^c) \\
& =-\log q(\mathbf{u}^c)-\log \left|\operatorname{det}\left(\frac{\partial f}{\partial \mathbf{F}^c}\right)\right|,
\end{align} 
where  $ \mathbf{u}^c = f(\mathbf{F}^c) $ and $ \mathbf{u}^c\sim\mathcal{N}(0, \mathbf{I}) $.

In contrast to prior NF implementations for generation \cite{yin2021graph} or human event AD \cite{hirschorn2023normalizing} which involve heavy network architectures, we draw the modeling idea from \cite{mao2021generating} by leveraging a lightweight constructed with a ten-layer MLP to ease training. We further adopt the QR decomposition to calculate the weights of MLPs and enforce the monotonic PReLU activation to ensure the invertibility of $f$. We next describe how to use the trained NF model to detect anomalous actions.

\noindent \textbf{Testing.} Generally, AD requires to endow the anomaly score to each testing sample for detection. Given the trained NF model $f^{*}$ on the normal action category, for an arbitrary test sample $\mathbf{Y}^{{c}_u}$, one straightforward scoring method is to feed it to our framework to yield the NLL as the anomaly score $S$. Despite the overall feasibility, we notice that such a scoring manner tends to be sensitive to semantically similar yet anomalous actions. This is because, since the same action category can cover diverse modes, the likelihood variation for normal samples can also vary noticeably among different action types. To nonetheless achieve a stabler detection performance, we propose a $K$-nearest neighbor (KNN)-based scoring approach to gain further robustness. Instead of the likelihood the NF finally outputs, inspired by image-based anomaly scoring strategies \cite{schlegl2017unsupervised}, we make use of the feature vector ${\mathbf{V} \in  \mathbb{R} ^ {P \times D_{V}} }$ regressed in the second layer from the last (i.e., one before the likelihood layer) to calculate the anomaly score. Specifically, we first feed the training samples in $\mathcal{X}^{c}$ and the testing sample $\mathbf{Y}^{{c}_u}$ to regress the feature vector set $\mathcal{V} = \{{\mathbf{V}^c_1, \cdots \mathbf{V}^c_N\} }$ and $\mathbf{V}^{c_u}$, respectively. We then search $K$ nearest vector neighbors to the query $\mathbf{V}^{c_u}$ from $\mathcal{V}$, and let the resulting distance set of the top $K$ samples to $\mathbf{V}^{c_u}$ be $\mathcal{D}=\{D_{1}, \cdots, D_{k}, \cdots, D_{K}\} $. Eventually, our anomaly score $S$ is derived by averaging the elements in $\mathcal{D}$, which is given by $S = \frac{1}{K}\sum_k D_{k}$. %The full anomaly scoring pipeline is shown in Alg. \ref{}.

\section{Experiment}
In this section, we report experimental results against baselines on two human action datasets to evaluate the effectiveness of our method to HAAD. We also show extensive ablative evaluations to gain deep insights into our model.

\noindent \textbf{Dataset.} Following previous human motion/action literature \cite{petrovich2021action} \cite{tevet2023human}, our evaluation is performed on the following two large-scale datasets: HumanAct12\cite{Guo_2020} and UESTC\cite{ji2019largescale}. 

\textbf{HumanAct12} \cite{Guo_2020} is derived from the PHSPD \cite{zou20203d} dataset as a subset with 1,191 motions composed of 90,099 frames. It is organized into 12 subjects where 12 types of actions with per-sequence annotation are included. The human pose in each frame is represented by 24 joints with 3D coordinates. We use the whole 12 actions in our experiment.

\textbf{UESTC} \cite{ji2019largescale} consists of 25K sequences in 118 subjects recorded with 8 static cameras. Compared to HumanAct12, the action annotation is performed more fine-grained, resulting in a total of 40 action categories. We manually select 10 actions in our experiments, leading to 6,000 more motion clips. The 25-joint and 6D-rotation configuration is adopted to represent the human poses.

\noindent \textbf{Implementation Details.} We train our model using the ADAM optimizer \cite{kingma2014adam} on RTX3090. We use a decaying learning rate of 0.001 at the start such that it evolves 1e-5 at the 50-th epoch. We train our model for 50 epochs for each action category. We set all the hidden size of GCNs to 128 and the number of graph convolution layer $L$ to 4. In KNN searching, the top 3 samples are used for efficiency. For HumanAct12, we divide the full body parts into 16 upper- and 8 lower-body joints, while for UESTC, the full body is decomposed to 17 upper- and 8 lower- joints. The first  DCT coefficients  $M$ are set to 10 and 5, respectively, on HumanAct12 and UESTC. 

\subsection{Evaluation}

\begin{table*}[]
\caption{\textbf{Quantitative results} against adapted baselines to HAAD on HumanAct12 with AUC. }
    \centering    \begin{tabular}{cccccccccccccc}
    \toprule
    \multirow{2}{*}{Method}& \multicolumn{12}{c}{HumanACT12}&\multirow{2}{*}{Avg.}\\
    \cmidrule(lr){2-13} 
         & \textit{Warm up}& \textit{Walk}& \textit{Run} & \textit{Jump} & \textit{Drink} & \textit{Lift dmbl}& \textit{Sit} & \textit{Eat} & \textit{Trn steer whl}& \textit{Phone} & \textit{Boxing} &\textit{Throw} &\\
         \hline
         STG-NF \cite{hirschorn2023normalizing}  & 0.720& 0.835& 0.568& 0.861& 0.751& 0.926& 0.878&  0.972& 0.855& \textbf{0.791}& 0.602&0.766&0.794\\
         MoCoDAD \cite{flaborea2023multimodal}& 0.664& 0.627& 0.516& 0.475& 0.631& 0.479& 0.406& 0.393& 0.480& 0.406&  0.358&0.497&0.494\\
         Our & \textbf{0.842}& \textbf{0.907}& \textbf{0.781}& \textbf{0.912}& \textbf{0.776}& \textbf{0.958}& \textbf{0.904}& \textbf{0.986}& \textbf{0.869}& 0.783 & \textbf{0.619}&\textbf{0.782}&\textbf{0.843}\\
         \bottomrule
    \end{tabular}

    \label{Table. HumanAct12 AUCs}
\end{table*}

\begin{table*}[]
\caption{\textbf{Quantitative results} against adapted baselines to HAAD on UESTC with AUC. Refer to Fig. \ref{fig:DCT_result} for full action label names. }
\centering  \begin{tabular}{cccccccccccc}
    %\addlinespace[3mm] 
    %\hdashline
    \toprule
    \multirow{2}{*}{Method}& \multicolumn{10}{c}{UESTC}&\multirow{2}{*}{Avg.}\\
    \cmidrule(lr){2-11} 
         & \textit{Punch}& \textit{Sng dbl rais}& \textit{Hd ackws crcl}& \textit{Std rtt}& \textit{Jump jack}& \textit{Kne to chst}& \textit{Rp skp}& \textit{Hgh knes run}& \textit{Squat}& \textit{Lft kck}&\\
         \hline
         STG-NF \cite{hirschorn2023normalizing}  & 0.832& 0.913& 0.827& 0.948
& \textbf{0.950}
& \textbf{0.946}&  0.974&  0.952
& 0.929& 0.966
&0.924\\
         MoCoDAD \cite{flaborea2023multimodal}& 0.436& 0.445& 0.503& 0.513
& 0.465
& 0.497& 0.512& 0.546
& 0.475& 0.508
&0.490\\
         Our & \textbf{0.918}& \textbf{0.967}& \textbf{0.948}& \textbf{0.950}& 0.941& 0.935& \textbf{0.976}& \textbf{0.960}& \textbf{0.933}& \textbf{0.973}&\textbf{0.950}\\
         \bottomrule
    \end{tabular}
    \label{Table. UESTC AUCs}
\end{table*}

We here report the results on the two datasets again prior action AD techniques. Since there is no prior work that tackles the task we introduce, we adapt the state-of-the-art pose-based video AD methods, STG-NF \cite{hirschorn2023normalizing} and MoCoDAD \cite{flaborea2023multimodal}, to our task. Specifically, STG-NF \cite{hirschorn2023normalizing} utilizes a frame-wise GCN to encode 2D pose information. We adapt their GCN-based spatial-temporal pose encoding module to take 3D coordinates or 6D rotation poses information as input. Similarly, we retrain the diffusion model of MoCoDAD \cite{flaborea2023multimodal} to learn from 3D or 6D representations. 

\noindent	\textbf{Evaluation Metrics.} We follow previous AD approaches \cite{hirschorn2023normalizing} \cite{markovitz2020graph} by adopting the Area Under the Receiver Operating Characteristic (ROC) curve (AUC) for quantitative evaluation. In particular, the ROC curve analyzes the relationship between the True Positive Rate (TPR,
$\overline{TP}/ (\overline{TP}+\overline{FN})$) and False Positive Rate (FPR, $\overline{FP}/ (\overline{FP}+\overline{TN})$) under a series of thresholds, where $\overline{TP}, \overline{FN}, \overline{FP}, \overline{TN}$ refer to the number of true positive, false negative, false positive, and true negative action samples, respectively. Then, the AUC metric is obtained via summing the area under such an ROC curve, with a larger AUC indicating stronger detection capacity.

%for the given testing action sequences, the corresponding anomaly scores and ground-truth action labels are first used to calculate the True Positive Rate (TPR) and False Positive Rate (FPR). 
%The ROC curve provides a visual representation of the trade-off between TPR and FPR with the threshold. The AUROC then measures the total area under such a ROC curve, and a large AUC is an indication of higher detection capacity.

\noindent{\textbf{Quantitative Results.}} We first provide the quantitative evaluation for anomaly detection performance. The results are summarized in 
Tabs. \ref{Table. HumanAct12 AUCs} and \ref{Table. UESTC AUCs}. For each column, all the results by ours and the compared models are retrained using the corresponding action selected as the normal class only. It can be confirmed that our method generally outperforms the compared baselines in all categories on both datasets. Specifically, our method outperforms MoCoDAD \cite{flaborea2023multimodal} by a large margin on all actions. Although MoCoDAD \cite{flaborea2023multimodal} is designed to characterize anomalies by measuring the multimodal prediction error, handling a large number of different action types within the testing samples can still be challenging. 

Let us now focus on STG-NF \cite{hirschorn2023normalizing}, which is also an NF-based AD for human activities. We can observe in Tabs. \ref{Table. HumanAct12 AUCs} and \ref{Table. UESTC AUCs} that on both datasets, STG-NF \cite{hirschorn2023normalizing} achieves comparable detection performance to ours. On general, the results on UESTC are more competitive to ours than those on HumanAct12. The reason is that, because UESTC is less nosier than HumanAct12, it provides an easier configuration for action AD. However, on some actions with close semantic behaviors, STG-NF can be less powerful to distinguish them. For example, as for \textit{Walk} and  \textit{Run} on HumanAct12, since the actions of these two categories are inherently similar, naively learning the action data with the NF formulation cannot model the inner difference between two actions well. This can be caused by the lack of awareness of body subsets during detection. Similar results can be further verified in  the actions of \textit{Head anticlockwise circling} and \textit{Standing rotation} on UESTC. By contrast, our method incorporates a multi-level partial body learning pipeline 
to characterize subtle anomalies within local body subsets, which yields higher detection accuracy for challenging anomaly types.

\begin{figure}[]
    \centering
    \includegraphics[scale=0.3]{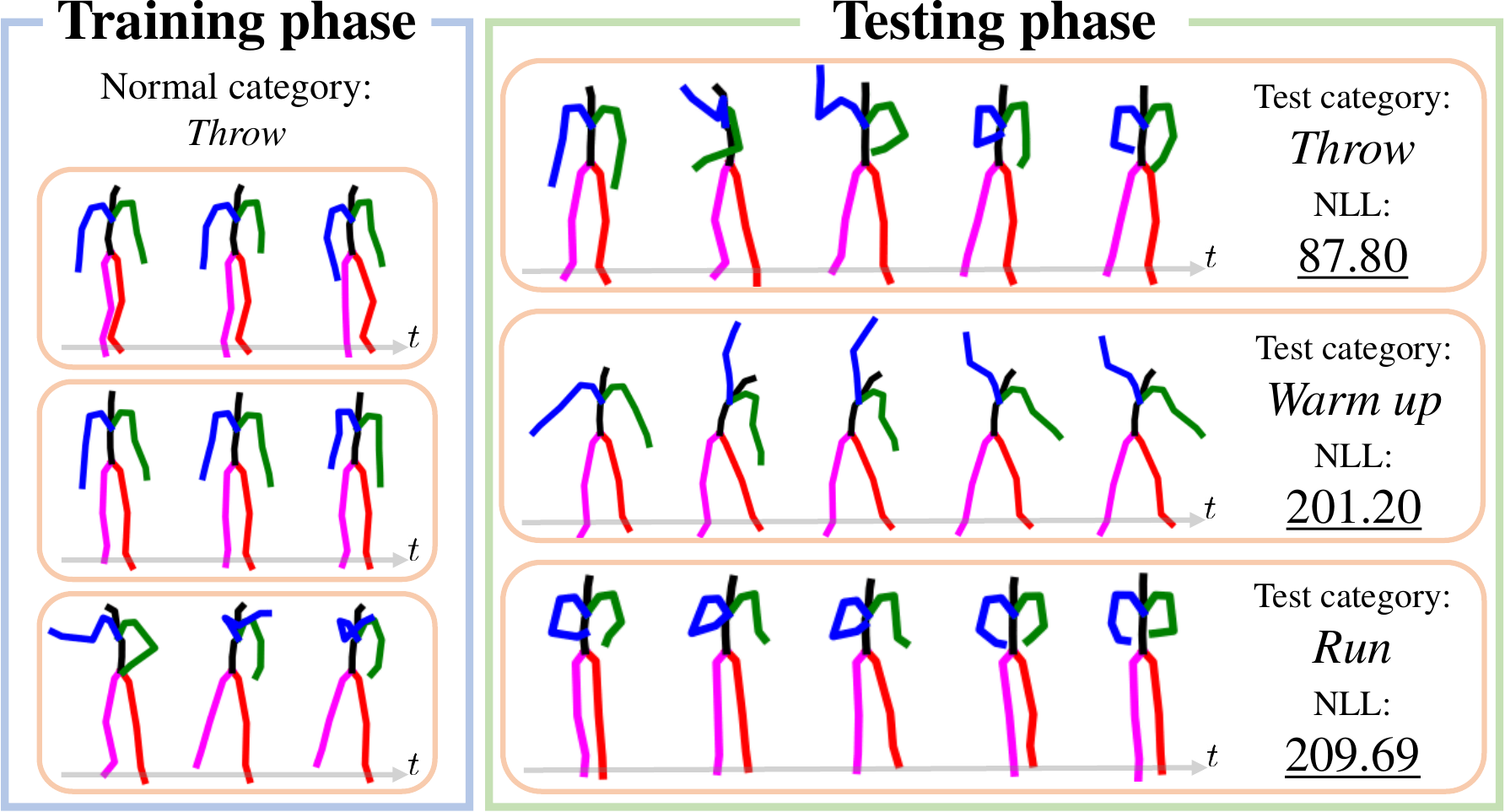}
    \caption{\textbf{Quantities results} by selecting \textit{Throw} in HumanAct12 as the normal action for training. }
    \label{fig:result_HA12_skltn}
\end{figure}
\begin{figure}[]
    \centering
    \includegraphics[scale=0.3]{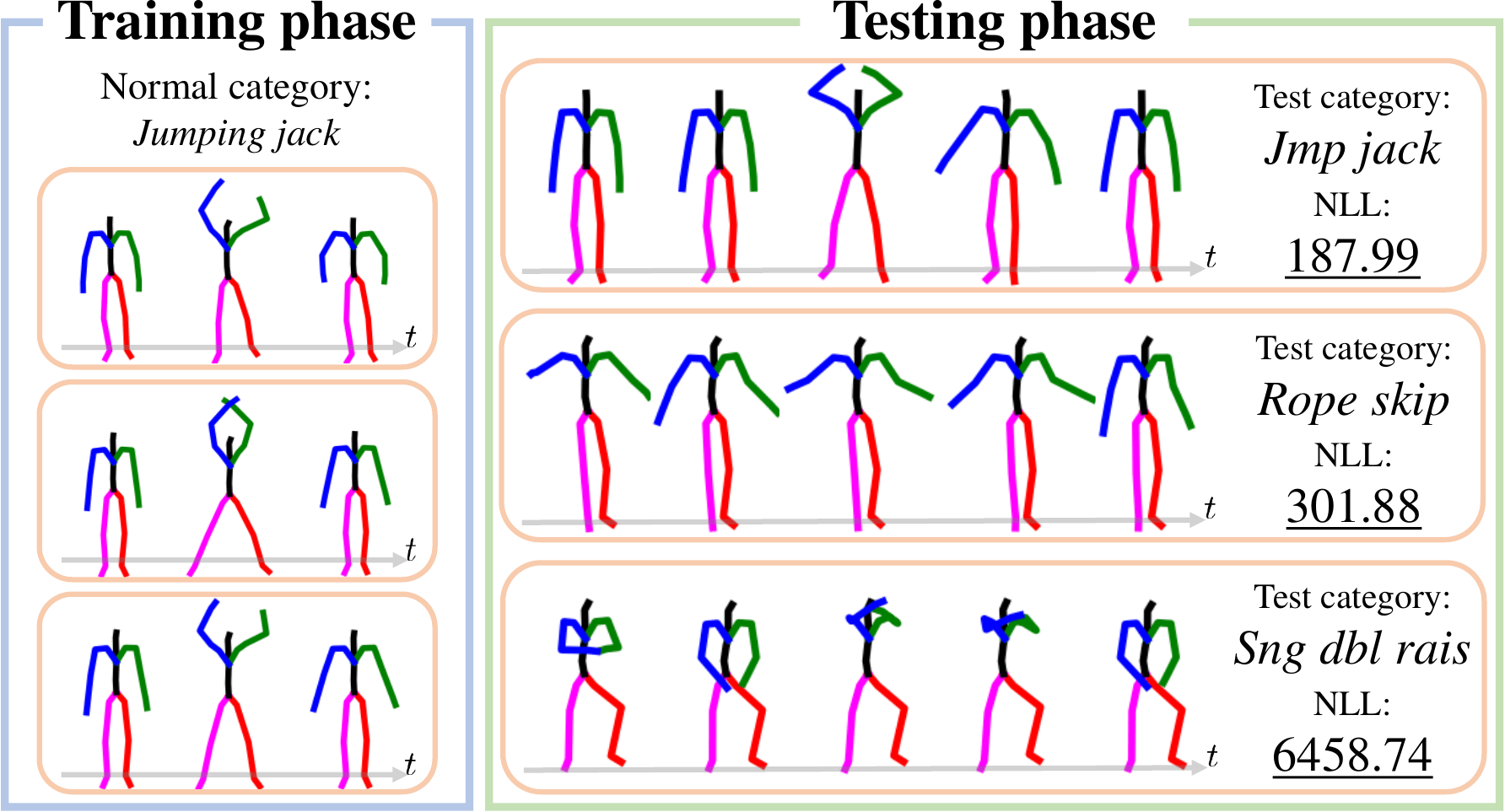}
    \caption{\textbf{Quantities results} by selecting \textit{Jumping jack} in UESTC as the normal action for training. }
    \label{fig:result_UESTC_skltn}
\end{figure}

\begin{table*}[]
\caption{ \textbf{Ablation studies on different encoding structures} on HumanAct12 with AUC.  }
    \centering    \begin{tabular}{cccccccccccccc}
    \toprule
    \multirow{2}{*}{Archtecture}& \multicolumn{12}{c}{HumanAct12}&\multirow{2}{*}{Avg.}\\
    \cmidrule(lr){2-13} 
         & \textit{Warm up}& \textit{Walk}& \textit{Run} & \textit{Jump} & \textit{Drink} & \textit{Lift dmbl}& \textit{Sit} & \textit{Eat} & \textit{Trn steer whl}& \textit{Phone} & \textit{Boxing} &\textit{Throw} &\\
         \hline
         GCN& 0.794& 0.837& 0.752& 0.891& 0.783& 0.949& 0.866&  0.964& 0.862& 0.816& \textbf{0.676}&0.732& 0.827\\
         GRU& 0.812& 0.780& \textbf{0.796}& \textbf{0.916}& 0.775& 0.951& 0.818& 0.983& 0.846& \textbf{0.848}&  0.611&0.677&0.818\\
 Transformer& 0.456& 0.791& 0.601& 0.574& 0.651& 0.475& 0.490& 0.467& 0.478& 0.458& 0.561& 0.542&0.545\\
         Ours (GCN \& DCT)& \textbf{0.842}& \textbf{0.907}& 0.781& 0.912& \textbf{0.776}& \textbf{0.952}& \textbf{0.904}& \textbf{0.986}& \textbf{0.869}& 0.783& 0.619&\textbf{0.782}&\textbf{0.843}\\
         \bottomrule
    \end{tabular}
    
    \label{Table. HumanAct12 ablation}
\end{table*}

\begin{table*}[]
\caption{\textbf{Ablation studies on different encoding structures} on UESTC with AUC. }
    \centering    \begin{tabular}{cccccccccccc}
    \toprule
    \multirow{2}{*}{Archtecture}& \multicolumn{10}{c}{UESTC}&\multirow{2}{*}{Avg.}\\
    \cmidrule(lr){2-11}& \textit{Punch}& \textit{Sng dbl rais}& \textit{Hd ackws crcl}& \textit{Std rtt}& \textit{Jump jack}& \textit{Kne to chst}& \textit{Rp skp}& \textit{Hgh knes run}& \textit{Squat}& \textit{Lft kck}&\\
    \hline
    GCN& 0.867& 0.944& \textbf{0.955}& 0.917
& 0.901
& 0.899& 0.941&  0.931
& 0.897&  0.959
& 0.921\\
         GRU& \textbf{0.953}& 0.966& 0.931& \textbf{0.978}
& \textbf{0.945}
& 0.887& \textbf{0.981}& \textbf{0.971}
& 0.893& 0.956
&0.946\\
 Transformer& 0.752& 0.776& 0.253& 0.229
& 0.500
& 0.770& 0.751& 0.736
& 0.509& 0.239
&0.551\\
         Ours (GCN \& DCT)& 0.918& \textbf{0.967}& 0.948& 0.950& 0.941& \textbf{0.935}& 0.976& 0.960& \textbf{0.933}& \textbf{0.973}&\textbf{0.950}\\
         \bottomrule
    \end{tabular}
    
    \label{Table. UESTC ablation}
\end{table*}
\noindent{\textbf{Quantitative Results.}} To provide deeper insights into our approach to HAAD,  we show qualitative detection results on both datasets in Figs. \ref{fig:result_HA12_skltn} and \ref{fig:result_UESTC_skltn}, respectively. Given the selected normal action for training in the left, we visualize three example testing sample actions in the right with the corresponding NLL score. It can be seen that the more different the actions are, the higher the NLL scores become. Specifically, because the training action \textit{throw} involves primarily the hand movements (Fig. \ref{fig:result_HA12_skltn}), the actions with strong leg dynamics causes a large NLL to indicate anomalies. This can be again better confirmed in Fig. \ref{fig:result_UESTC_skltn} on the UESTC dataset. When the jumping and hand-clapping motions constitute the moving trends of training samples, the seated testing sample (last row in Fig. \ref{fig:result_UESTC_skltn}, right), which shows significant action disparity with training data, induces a remarkably great NLL (i.e., 6458.74). We expect this to be due to that our multi-level framework contributes to the strength in discovering locally inconsistent action anomaly patterns. Also, for both datasets, the training samples with the normal categories lead to low NLL scores. We can thus verify the feasibility and effectiveness of our NF-based formulation for the task of HAAD.

\subsection{Ablation Study}
To gain more understanding of our method, we perform the following ablative evaluations to examine the role of each component in our model.

\begin{figure*}[]
    \centering
    \includegraphics[width=1\linewidth]{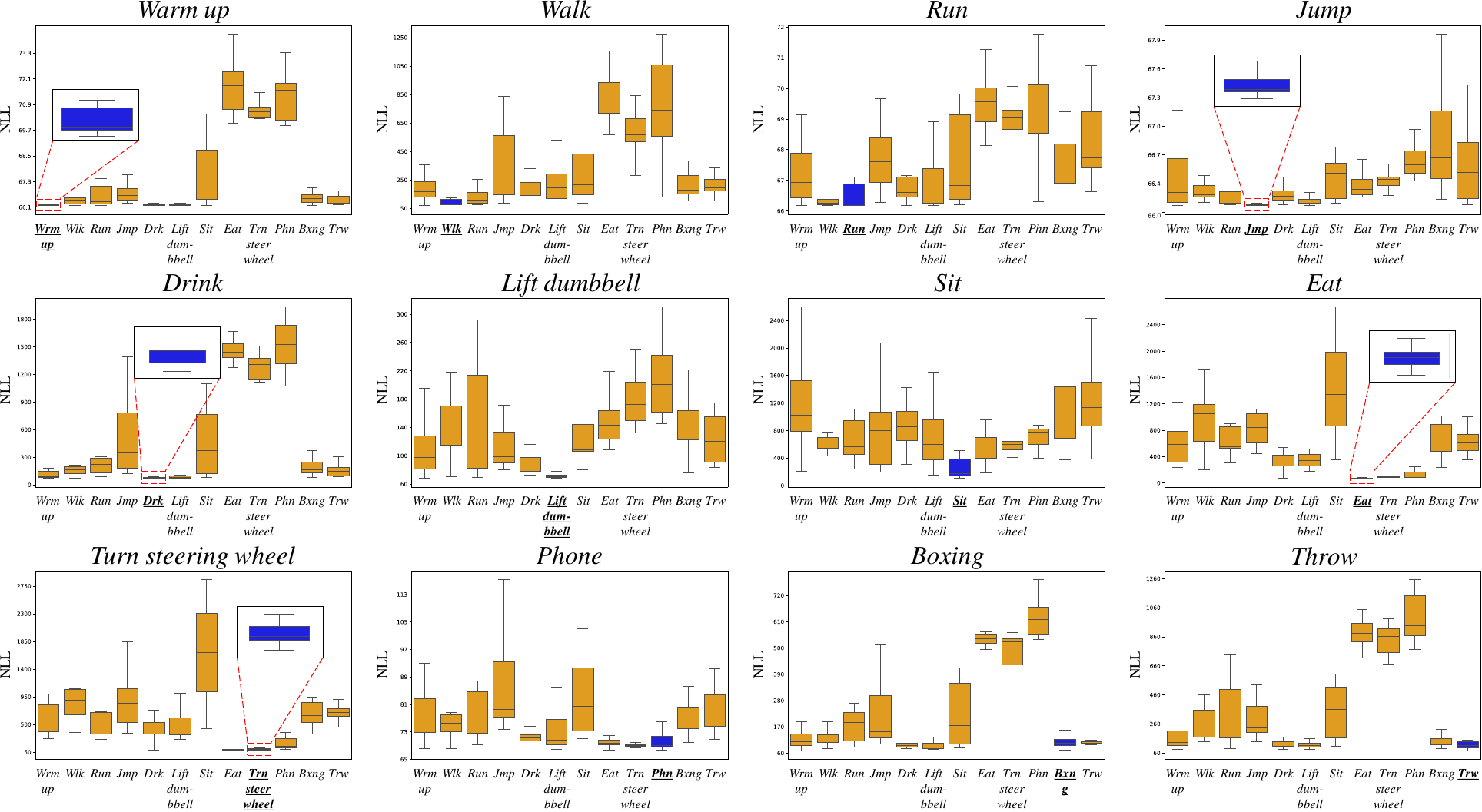}
    \caption{\textbf{Box-and-whisker plot} for testing phase NLL on HumanAct12. The action label shown on top of each diagram refers to the corresponding selected normal action class (colored in blue). The anomalous actions are shown in orange boxes.} %The lower the blue box is below the other boxes, the better the classification results.}
    \label{fig:NLL_score_HA12}
\end{figure*}

\begin{figure*}[]
    \centering
    \includegraphics[width=1\linewidth]{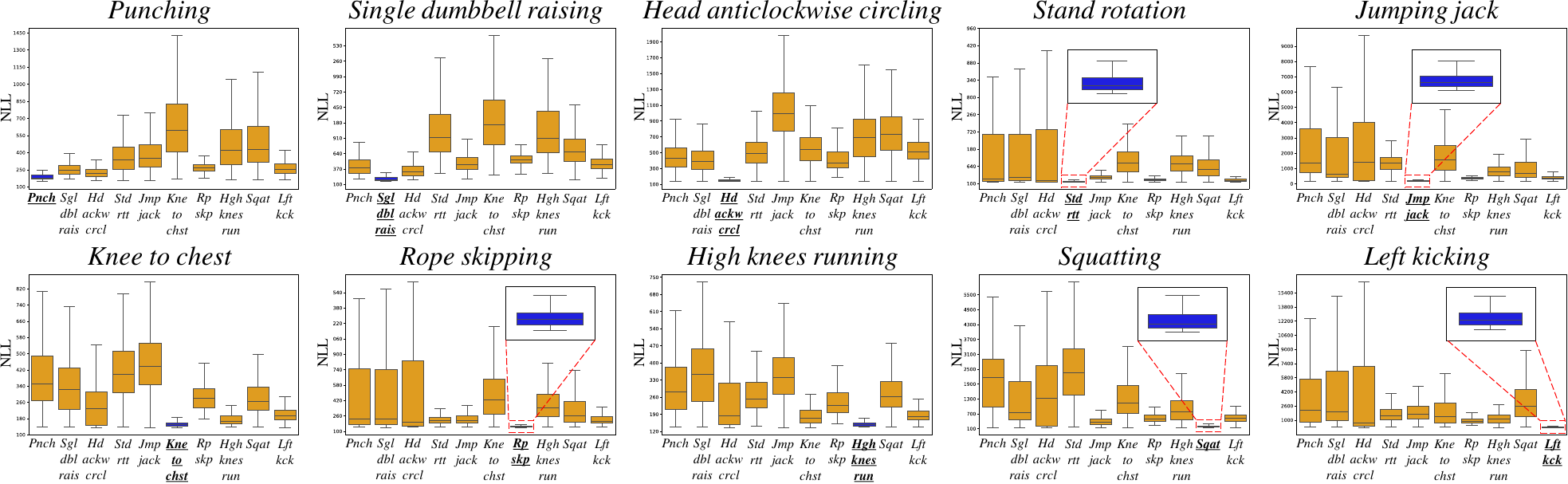}
    \caption{\textbf{Box-and-whisker plot} for testing phase NLL on UESTC. The action label shown on top of each diagram refers to the corresponding selected normal action class (colored in blue). The anomalous actions are shown in orange boxes.}
    \label{fig:NLL_score_UESTC}
\end{figure*}

\noindent{\textbf{Encoding Techniques.}}
Our model leverages GCN to encode the human motion in the frequency domain for further NF learning. To investigate the validity, we remove the DCT module and prepare two possible temporal encoding architectures, Transformer \cite{vaswani2017attention} and Gated Recurrent Unit (GRU) \cite{cho2014learning}, to substitute the GCN during training. The results are shown in Tabs. \ref{Table. HumanAct12 ablation} and \ref{Table. UESTC ablation}. It can be observed that on both datasets, GRU achieves comparable performance to GCN in detection accuracy, outperforming the Transformer by a large margin. We assume this is because since the training data for each scenario is limited to a single action type, it can be difficult to train the Transformer from scratch considering its data-hungry attribute. Hence, Transformer-based models are not suitable for the task of HAAD. As for the GCN and GRU, we can see from the bottom rows in  Tabs. \ref{Table. HumanAct12 ablation} and \ref{Table. UESTC ablation} that with the introduction of DCT encoding to learn in the frequency domain, our GCN achieves marginal improvements compared to the GRU. Since the recorded motions can be jittery, the DCT smoothing plays a positive role in stabilizing the data to promote feature embedding. Note that GRU and Transformer are devised to model sequential data and thus cannot be straightforwardly applied in the frequency domain. The above analysis verifies the significance of introducing GCN and DCT jointly in the feature extraction phase prior to anomaly scoring. 

\begin{table*}[]
\caption{ \textbf{Ablation studies} on multi-level action feature learning by employing different body subsets in our framework with AUC on HumanAct12.  }
    \centering    \begin{tabular}{cllccccccccccccc}
    \toprule
    \multirow{2}{*}{Full}& \multirow{2}{*}{Up}&\multirow{2}{*}{Low}& \multicolumn{12}{c}{HumanAct12}&\multirow{2}{*}{Avg.}\\
    \cmidrule(lr){4-15}& && \textit{Warm up}& \textit{Walk}& \textit{Run} & \textit{Jump} & \textit{Drink} & \textit{Lift dmbl}& \textit{Sit} & \textit{Eat} & \textit{Trn steer whl}& \textit{Phone} & \textit{Boxing} &\textit{Throw} &\\
         \hline
         $\checkmark$&           &          & 0.749& 0.828& 0.668& 0.693& \textbf{0.789}& 0.943& 0.831&  0.977& 0.860& \textbf{0.790}& 0.524&0.703& 0.779\\
 $\checkmark$& $\checkmark$& & 0.765& 0.820& 0.721& 0.786& 0.760& 0.950& 0.846& 0.969& 0.815& 0.681& 0.530& 0.686&0.777\\
         $\checkmark$& &$\checkmark$& 0.740& 0.819& 0.688& 0.629& 0.770& 0.933& 0.726& 0.972& 0.823& 0.722&  0.506&0.636&0.747\\
 $\checkmark$& $\checkmark$&$\checkmark$& \textbf{0.842}& \textbf{0.907}& \textbf{0.781}& \textbf{0.912}& 0.776& \textbf{0.952}& \textbf{0.904}& \textbf{0.986}& \textbf{0.869}& 0.783& \textbf{0.619}& \textbf{0.782}&\textbf{0.843}\\
    \bottomrule
    \end{tabular}
    
    \label{Table. HumanAct12 multi-level}
\end{table*}

\begin{table*}[]
\caption{ \textbf{Ablation studies} on multi-level action feature learning by employing different body subsets in our framework with AUC on UESTC. }
    \centering    \begin{tabular}{cllccccccccccc}
    \toprule
    \multirow{2}{*}{Full}& \multirow{2}{*}{Up}&\multirow{2}{*}{Low}
& \multicolumn{10}{c}{UESTC}&\multirow{2}{*}{Avg.}\\
    \cmidrule(lr){4-13}& &
& \textit{Punch}& \textit{Sng dbl rais}& \textit{Hd ackws crcl}& \textit{Std rtt}& \textit{Jump jack}& \textit{Kne to chst}& \textit{Rp skp}& \textit{Hgh knes run}& \textit{Squat}& \textit{Lft kck}&\\
         \hline
         $\checkmark$& &
& 0.892& 0.963& \underline{0.948}& 0.928& 0.927& 0.926& 0.964&  0.949& 0.920&  0.960& 0.938\\
         $\checkmark$& $\checkmark$&
& \textbf{0.920}& \textbf{0.976}& 0.946& \textbf{0.955}& \underline{0.940}& \textbf{0.941}& \underline{0.975}& \underline{0.958}& \textbf{0.936}& \underline{0.970}&\textbf{0.951}\\
 $\checkmark$& &$\checkmark$
& 0.906& 0.966& \textbf{0.949}& \underline{0.951}& 0.937& 0.929& \textbf{0.976}& \underline{0.958}& 0.930& \textbf{0.973}&0.947\\
         $\checkmark$& $\checkmark$&$\checkmark$& \underline{0.918}& \underline{0.967}& \underline{0.948}& 0.950& \textbf{0.941}& \underline{0.935}& \textbf{0.976}& \textbf{0.960}& \underline{0.933}& \textbf{0.973}&\underline{0.950}\\
         \bottomrule
    \end{tabular}
    
    \label{Table. UESTC multi-level}
\end{table*}

\noindent{\textbf{NLL for Anomaly Modeling.}} Our key insight for HAAD is by quantifying the anomalous degree with NLL. In addition to the quantitative results in Tabs. \ref{Table. HumanAct12 AUCs} and \ref{Table. UESTC AUCs}, we further provide more results in Figs. \ref{fig:NLL_score_HA12} and \ref{fig:NLL_score_UESTC} to study whether such a modeling strategy. In Figs. \ref{fig:NLL_score_HA12} and \ref{fig:NLL_score_UESTC}, we visualize the distribution of NLL as a box-and-whisker plot. We can see that the testing actions selected as normal generally induce a noticeably low NLL scores on both datasets, which validates our motivation of formulating the AD framework with NF. However, we notice that for the actions with similar motion patterns, the NLL scores tend to be close. For example, when \textit{Drink} in HumanAct12 is considered normal (Fig. \ref{fig:NLL_score_HA12}), the actions of  \textit{Lift dumbbell} and  \textit{Throw} also induce comparably low NLL with \textit{Drink}. It is interesting to point out that these three actions share a common movement in raising his/her hand near the face while leaving other parts nearly still. Even for these challenging cases, our method still achieves decent detection accuracy to distinguish the anomalies (Tab, \ref{Table. HumanAct12 AUCs}\ref{Table. UESTC AUCs}). We can thus verify the effectiveness of using NLL for HAAD.

\noindent{\textbf{Multi-level Feature Fusion.}} To examine the superiority of our multi-level manner of action learning, we here ablate each portion of body subsets in our model training and summarize the results on Tabs. \ref{Table. HumanAct12 multi-level} and \ref{Table. UESTC multi-level} for both datasets. As can be confirmed, using full body and each portion of the body subset result in a noticeable accuracy gain on HumanAct12, whereas on UESTC, involving the body subsets (either upper- or lower-body portions or both) generally outperforms the scenario where only the full-body is utilized. We notice that, for some actions with typical partial body movements, such as \textit{Punch} or \textit{Left kick} on UESTC, introducing the corresponding upper or lower bodies only lead to larger performance improvement. We assume the reason to be that compared to HumanAct12 in which the samples categories include an equal proportion of characteristic actions for both body parts, UESTC is constituted by more active upper-body movements. Therefore, the results in Tabs. \ref{Table. HumanAct12 multi-level} and \ref{Table. UESTC multi-level} evidence that introducing the multi-level mechanism ensures a better modeling of local action patterns to improve the detection accuracy. Moreover, it is worth mentioning an interesting direction in exploring adaptive or weighted body subset learning approach, which we would like to resolve in the future.

\noindent{\textbf{Anomaly Scoring Schemes.}} Different from  STG-NF \cite{hirschorn2023normalizing} which directly exploits the NLL, the anomalies are scored based on the KNN in our implementation. To investigate the underlying effect, we compare in Tab. \ref{Table. HumanAct12 anomaly score} the detection accuracy of these two different scoring schemes. Overall, our proposed KNN-based scoring achieves higher average AUCs, especially on the UESTC dataset. The reason can be attributed two-fold: (i) During the NF modeling optimizing the NLL, the feature vector also gradually learns to embed quality action latents; (ii) Since the KNN returns multiple neighboring vectors and our scheme averages over them to score, the negative influence imposed by the outliers can be mitigated to stabilize the detection performance. In particular, because UESTC covers a more diverse motion modes than HumanAct12 per action category with more potential outliers, (ii) also explains why our scoring method performs more effectively on UESTC.

\begin{figure*}[]
    \begin{minipage}[b]{0.47\linewidth}
    \centering
    \includegraphics[width=0.8\linewidth]{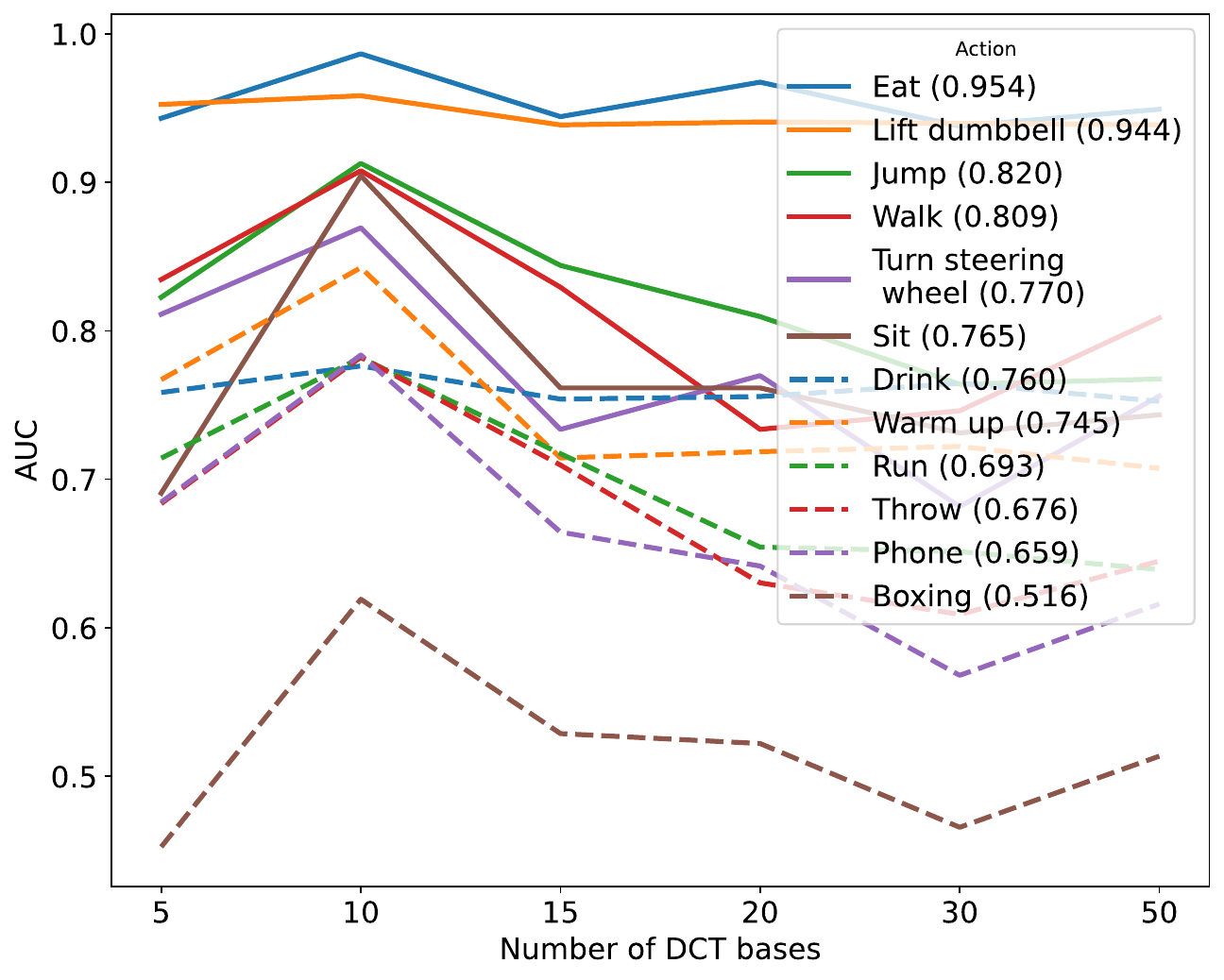}
    \end{minipage}
    \begin{minipage}[b]{0.47\linewidth}
         \centering
    \includegraphics[width=0.8\linewidth]{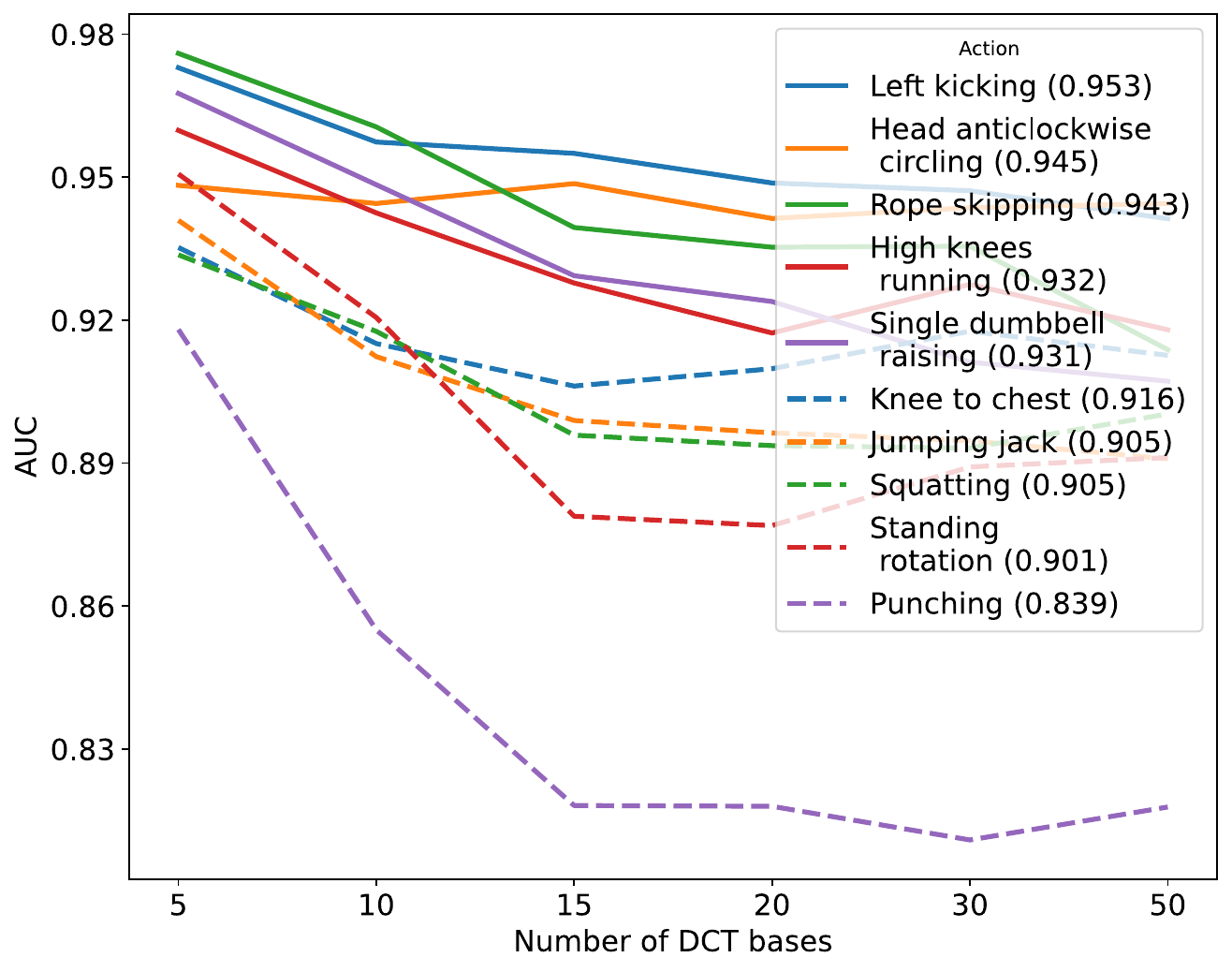}
    \end{minipage}
    \caption{\textbf{Ablation studies} on the number of DCT coefficients $M$ via AUC on HumanAct12 (left) and UESTC (right). }
    \label{fig:DCT_result}
\end{figure*}

\begin{table}[]
\caption{ \textbf{Ablation studies} on the anomaly scoring schemes on both datasets with average AUC. NLL refers to directly using NLL for scoring, while Feature-KNN denotes our proposed KNN based scoring with feature vectors. }
    \centering    \begin{tabular}{lcc}
    \toprule
     & HumanAct12&UESTC\\
         \hline
                              NLL & 0.842&0.931\\
   Feature-KNN (Ours)& \textbf{0.843}& \textbf{0.950}\\
   \bottomrule
    \end{tabular}
    
    \label{Table. HumanAct12 anomaly score}
\end{table}

%\noindent{\textbf{Multi-level Feature Fusion.}}Here, we provide the performance of our method with different using body parts. We focus on more detailed movements by dividing the human skeleton into multi-level body parts and extracting features. Tab. \ref{Table. HumanAct12 multi-level} and \ref{Table. UESTC multi-level} shows the results of evaluation by different body parts, utilizing three types of body parts: whole body, upper body, and lower body. In Tab. \ref{Table. HumanAct12 multi-level}, we confirm that in most categories, the model with the three body parts outperforms the other models. Focusing on walking and running categories, which are similar movements, the AUC values of the model with three body parts are improved compared to the full body parts model. This is because the combination of the detailed features of the body parts allowed the model to capture different fine movements, even if the movements were similar from the aspect of the whole body. In Tab. \ref{Table. UESTC multi-level}, the Full and Up body models outperform the other models for most of the movement labels, however, there is almost no difference in AUC values compared to the three body part models, and the three body parts model has the second-highest AUC average values. In addition, Tab. \ref{Table. UESTC multi-level} also shows that the three body parts model has a higher AUC value than the model with the full body.

\noindent{\textbf{The Number of DCT Coefficients.}}
To study the optimal number of DCT coefficients $M$ during learning, we present in Fig. \ref{fig:DCT_result} the detection results of each action category on both datasets by varying $M$. Overall, a smaller $M$ suggests a stronger smoothing intensity, while a larger $M$ means a better preservation of the high-frequency motions. We can see in Fig. \ref{fig:DCT_result} that on both datasets, a small DCT setting contributes to higher AUC scores on all categories. Considering the noise and jitter within the sequential action data, a smaller $M$ enables mitigating the resulting negative influence that hinders learning quality motion latents. Based on the results, we set $M$ to 10 and 5 on HumanAct12 and UESTC, respectively, to ensure a satisfactory detection performance.

\section{Conclusion}
We have proposed a new task, human action anomaly detection (HAAD), which aims to detect anomalous 3D motion patterns with a target normal category of human action in an unsupervised manner. It differs from previous human activity AD tasks in that it introduces the exact semantic action type to indicate the normal or anomalous data category. To address this, we propose handling HAAD in the frequency domain by performing DCT on the input temporal motion sample. We incorporate multi-level branches by separating the human body into several subsets such that our model can also locally characterize the anomaly within some specific body portions. Moreover, we propose a KNN-based anomaly scoring scheme to gain further robustness to motion outliers during testing. Extensive experimental results and ablative evaluations on two large-scale human motion datasets demonstrate that our method outperforms other baseline approaches constructed by extending state-of-the-art human event AD models to our task.

Despite the satisfactory detection efficiency, we notice that instead of employing the features in both of the multi-level branch, 
for some specific actions, selectively adopting one body portion or weighting the features prior to fusion may lead to better performance. Also, a learnable design of DCT coefficient number can be beneficial in smoothing. We would like to explore these two interesting future directions.

% or


%\appendices

%\section{}

%\section*{Acknowledgment}


\ifCLASSOPTIONcaptionsoff
  \newpage
\fi



%\bibliographystyle{IEEEtran}

\bibliographystyle{abbrv}
\bibliography{reference}

\begin{thebibliography}{10}

\bibitem{bruce2022mmnet}
X.~Bruce, Y.~Liu, X.~Zhang, S.-h. Zhong, and K.~C. Chan.
\newblock Mmnet: A model-based multimodal network for human action recognition
  in rgb-d videos.
\newblock {\em IEEE Transactions on Pattern Analysis and Machine Intelligence},
  45(3):3522--3538, 2022.

\bibitem{caetano2019skelemotion}
C.~Caetano, J.~Sena, F.~Br{\'e}mond, J.~A. Dos~Santos, and W.~R. Schwartz.
\newblock Skelemotion: A new representation of skeleton joint sequences based
  on motion information for 3d action recognition.
\newblock In {\em 2019 16th IEEE international conference on advanced video and
  signal based surveillance (AVSS)}, pages 1--8. IEEE, 2019.

\bibitem{cho2014learning}
K.~Cho, B.~Van~Merri{\"e}nboer, C.~Gulcehre, D.~Bahdanau, F.~Bougares,
  H.~Schwenk, and Y.~Bengio.
\newblock Learning phrase representations using rnn encoder-decoder for
  statistical machine translation.
\newblock {\em arXiv preprint arXiv:1406.1078}, 2014.

\bibitem{cohen2020sub}
N.~Cohen and Y.~Hoshen.
\newblock Sub-image anomaly detection with deep pyramid correspondences.
\newblock {\em arXiv preprint arXiv:2005.02357}, 2020.

\bibitem{defard2021padim}
T.~Defard, A.~Setkov, A.~Loesch, and R.~Audigier.
\newblock Padim: a patch distribution modeling framework for anomaly detection
  and localization.
\newblock In {\em International Conference on Pattern Recognition}, pages
  475--489. Springer, 2021.

\bibitem{doshi2020continual}
K.~Doshi and Y.~Yilmaz.
\newblock Continual learning for anomaly detection in surveillance videos.
\newblock In {\em Proceedings of the IEEE/CVF conference on computer vision and
  pattern recognition workshops}, pages 254--255, 2020.

\bibitem{duan2022revisiting}
H.~Duan, Y.~Zhao, K.~Chen, D.~Lin, and B.~Dai.
\newblock Revisiting skeleton-based action recognition.
\newblock In {\em Proceedings of the IEEE/CVF conference on computer vision and
  pattern recognition}, pages 2969--2978, 2022.

\bibitem{feichtenhofer2016convolutional}
C.~Feichtenhofer, A.~Pinz, and A.~Zisserman.
\newblock Convolutional two-stream network fusion for video action recognition.
\newblock In {\em Proceedings of the IEEE conference on computer vision and
  pattern recognition}, pages 1933--1941, 2016.

\bibitem{flaborea2023multimodal}
A.~Flaborea, L.~Collorone, G.~M.~D. Di~Melendugno, S.~D'Arrigo, B.~Prenkaj, and
  F.~Galasso.
\newblock Multimodal motion conditioned diffusion model for skeleton-based
  video anomaly detection.
\newblock In {\em Proceedings of the IEEE/CVF International Conference on
  Computer Vision}, pages 10318--10329, 2023.

\bibitem{georgescu2021anomaly}
M.-I. Georgescu, A.~Barbalau, R.~T. Ionescu, F.~S. Khan, M.~Popescu, and
  M.~Shah.
\newblock Anomaly detection in video via self-supervised and multi-task
  learning.
\newblock In {\em Proceedings of the IEEE/CVF conference on computer vision and
  pattern recognition}, pages 12742--12752, 2021.

\bibitem{georgescu2021background}
M.~I. Georgescu, R.~T. Ionescu, F.~S. Khan, M.~Popescu, and M.~Shah.
\newblock A background-agnostic framework with adversarial training for
  abnormal event detection in video.
\newblock {\em IEEE transactions on pattern analysis and machine intelligence},
  44(9):4505--4523, 2021.

\bibitem{goodfellow2014generative}
I.~Goodfellow, J.~Pouget-Abadie, M.~Mirza, B.~Xu, D.~Warde-Farley, S.~Ozair,
  A.~Courville, and Y.~Bengio.
\newblock Generative adversarial nets.
\newblock {\em Advances in neural information processing systems}, 27, 2014.

\bibitem{Guo_2020}
C.~Guo, X.~Zuo, S.~Wang, S.~Zou, Q.~Sun, A.~Deng, M.~Gong, and L.~Cheng.
\newblock Action2motion.
\newblock In {\em Proceedings of the 28th {ACM} International Conference on
  Multimedia}. {ACM}, oct 2020.

\bibitem{hara2018can}
K.~Hara, H.~Kataoka, and Y.~Satoh.
\newblock Can spatiotemporal 3d cnns retrace the history of 2d cnns and
  imagenet?
\newblock In {\em Proceedings of the IEEE conference on Computer Vision and
  Pattern Recognition}, pages 6546--6555, 2018.

\bibitem{hirschorn2023normalizing}
O.~Hirschorn and S.~Avidan.
\newblock Normalizing flows for human pose anomaly detection.
\newblock In {\em Proceedings of the IEEE/CVF International Conference on
  Computer Vision}, pages 13545--13554, 2023.

\bibitem{ji2019largescale}
Y.~Ji, F.~Xu, Y.~Yang, F.~Shen, H.~T. Shen, and W.-S. Zheng.
\newblock A large-scale varying-view rgb-d action dataset for arbitrary-view
  human action recognition.
\newblock {\em arXiv preprint arXiv:1904.10681}, 2019.

\bibitem{jiang2011anomalous}
F.~Jiang, J.~Yuan, S.~A. Tsaftaris, and A.~K. Katsaggelos.
\newblock Anomalous video event detection using spatiotemporal context.
\newblock {\em Computer Vision and Image Understanding}, 115(3):323--333, 2011.

\bibitem{kim2023cross}
S.~Kim, D.~Ahn, and B.~C. Ko.
\newblock Cross-modal learning with 3d deformable attention for action
  recognition.
\newblock In {\em Proceedings of the IEEE/CVF International Conference on
  Computer Vision}, pages 10265--10275, 2023.

\bibitem{kingma2014adam}
D.~P. Kingma and J.~Ba.
\newblock Adam: A method for stochastic optimization.
\newblock {\em arXiv preprint arXiv:1412.6980}, 2014.

\bibitem{kingma2013auto}
D.~P. Kingma and M.~Welling.
\newblock Auto-encoding variational bayes.
\newblock {\em arXiv preprint arXiv:1312.6114}, 2013.

\bibitem{kipf2016semi}
T.~N. Kipf and M.~Welling.
\newblock Semi-supervised classification with graph convolutional networks.
\newblock {\em arXiv preprint arXiv:1609.02907}, 2016.

\bibitem{lee2017ensemble}
I.~Lee, D.~Kim, S.~Kang, and S.~Lee.
\newblock Ensemble deep learning for skeleton-based action recognition using
  temporal sliding lstm networks.
\newblock In {\em Proceedings of the IEEE international conference on computer
  vision}, pages 1012--1020, 2017.

\bibitem{li2019actional}
M.~Li, S.~Chen, X.~Chen, Y.~Zhang, Y.~Wang, and Q.~Tian.
\newblock Actional-structural graph convolutional networks for skeleton-based
  action recognition.
\newblock In {\em Proceedings of the IEEE/CVF conference on computer vision and
  pattern recognition}, pages 3595--3603, 2019.

\bibitem{li2013anomaly}
W.~Li, V.~Mahadevan, and N.~Vasconcelos.
\newblock Anomaly detection and localization in crowded scenes.
\newblock {\em IEEE transactions on pattern analysis and machine intelligence},
  36(1):18--32, 2013.

\bibitem{lu2018anomaly}
Y.~Lu and P.~Xu.
\newblock Anomaly detection for skin disease images using variational
  autoencoder.
\newblock {\em arXiv preprint arXiv:1807.01349}, 2018.

\bibitem{luo2021normal}
W.~Luo, W.~Liu, and S.~Gao.
\newblock Normal graph: Spatial temporal graph convolutional networks based
  prediction network for skeleton based video anomaly detection.
\newblock {\em Neurocomputing}, 444:332--337, 2021.

\bibitem{mao2021generating}
W.~Mao, M.~Liu, and M.~Salzmann.
\newblock Generating smooth pose sequences for diverse human motion prediction.
\newblock In {\em Proceedings of the IEEE/CVF International Conference on
  Computer Vision}, pages 13309--13318, 2021.

\bibitem{mao2019learning}
W.~Mao, M.~Liu, M.~Salzmann, and H.~Li.
\newblock Learning trajectory dependencies for human motion prediction.
\newblock In {\em Proceedings of the IEEE/CVF international conference on
  computer vision}, pages 9489--9497, 2019.

\bibitem{markovitz2020graph}
A.~Markovitz, G.~Sharir, I.~Friedman, L.~Zelnik-Manor, and S.~Avidan.
\newblock Graph embedded pose clustering for anomaly detection.
\newblock In {\em Proceedings of the IEEE/CVF Conference on Computer Vision and
  Pattern Recognition}, pages 10539--10547, 2020.

\bibitem{morais2019learning}
R.~Morais, V.~Le, T.~Tran, B.~Saha, M.~Mansour, and S.~Venkatesh.
\newblock Learning regularity in skeleton trajectories for anomaly detection in
  videos.
\newblock In {\em Proceedings of the IEEE/CVF conference on computer vision and
  pattern recognition}, pages 11996--12004, 2019.

\bibitem{pang2023skeleton}
C.~Pang, X.~Lu, and L.~Lyu.
\newblock Skeleton-based action recognition through contrasting two-stream
  spatial-temporal networks.
\newblock {\em IEEE Transactions on Multimedia}, 2023.

\bibitem{park2020learning}
H.~Park, J.~Noh, and B.~Ham.
\newblock Learning memory-guided normality for anomaly detection.
\newblock In {\em Proceedings of the IEEE/CVF conference on computer vision and
  pattern recognition}, pages 14372--14381, 2020.

\bibitem{petrovich2021action}
M.~Petrovich, M.~J. Black, and G.~Varol.
\newblock Action-conditioned 3d human motion synthesis with transformer vae.
\newblock In {\em Proceedings of the IEEE/CVF International Conference on
  Computer Vision}, pages 10985--10995, 2021.

\bibitem{reiss2021panda}
T.~Reiss, N.~Cohen, L.~Bergman, and Y.~Hoshen.
\newblock Panda: Adapting pretrained features for anomaly detection and
  segmentation.
\newblock In {\em Proceedings of the IEEE/CVF Conference on Computer Vision and
  Pattern Recognition}, pages 2806--2814, 2021.

\bibitem{ristea2022self}
N.-C. Ristea, N.~Madan, R.~T. Ionescu, K.~Nasrollahi, F.~S. Khan, T.~B.
  Moeslund, and M.~Shah.
\newblock Self-supervised predictive convolutional attentive block for anomaly
  detection.
\newblock In {\em Proceedings of the IEEE/CVF conference on computer vision and
  pattern recognition}, pages 13576--13586, 2022.

\bibitem{rodrigues2020multi}
R.~Rodrigues, N.~Bhargava, R.~Velmurugan, and S.~Chaudhuri.
\newblock Multi-timescale trajectory prediction for abnormal human activity
  detection.
\newblock In {\em Proceedings of the IEEE/CVF Winter Conference on Applications
  of Computer Vision}, pages 2626--2634, 2020.

\bibitem{roth2022towards}
K.~Roth, L.~Pemula, J.~Zepeda, B.~Sch{\"o}lkopf, T.~Brox, and P.~Gehler.
\newblock Towards total recall in industrial anomaly detection.
\newblock In {\em Proceedings of the IEEE/CVF Conference on Computer Vision and
  Pattern Recognition}, pages 14318--14328, 2022.

\bibitem{sabokrou2017deep}
M.~Sabokrou, M.~Fayyaz, M.~Fathy, and R.~Klette.
\newblock Deep-cascade: Cascading 3d deep neural networks for fast anomaly
  detection and localization in crowded scenes.
\newblock {\em IEEE Transactions on Image Processing}, 26(4):1992--2004, 2017.

\bibitem{schlegl2017unsupervised}
T.~Schlegl, P.~Seeb{\"o}ck, S.~M. Waldstein, U.~Schmidt-Erfurth, and G.~Langs.
\newblock Unsupervised anomaly detection with generative adversarial networks
  to guide marker discovery.
\newblock In {\em International conference on information processing in medical
  imaging}, pages 146--157. Springer, 2017.

\bibitem{shuchang2022survey}
Z.~Shuchang.
\newblock A survey on human action recognition.
\newblock {\em arXiv preprint arXiv:2301.06082}, 2022.

\bibitem{tevet2023human}
G.~Tevet, S.~Raab, B.~Gordon, Y.~Shafir, D.~Cohen-or, and A.~H. Bermano.
\newblock Human motion diffusion model.
\newblock In {\em The Eleventh International Conference on Learning
  Representations}, 2023.

\bibitem{vaswani2017attention}
A.~Vaswani, N.~Shazeer, N.~Parmar, J.~Uszkoreit, L.~Jones, A.~N. Gomez,
  {\L}.~Kaiser, and I.~Polosukhin.
\newblock Attention is all you need.
\newblock {\em Advances in neural information processing systems}, 30, 2017.

\bibitem{wang2021student}
G.~Wang, S.~Han, E.~Ding, and D.~Huang.
\newblock Student-teacher feature pyramid matching for anomaly detection.
\newblock {\em arXiv preprint arXiv:2103.04257}, 2021.

\bibitem{wang2022video}
G.~Wang, Y.~Wang, J.~Qin, D.~Zhang, X.~Bao, and D.~Huang.
\newblock Video anomaly detection by solving decoupled spatio-temporal jigsaw
  puzzles.
\newblock In {\em European Conference on Computer Vision}, pages 494--511.
  Springer, 2022.

\bibitem{wang2018video}
S.~Wang, E.~Zhu, J.~Yin, and F.~Porikli.
\newblock Video anomaly detection and localization by local motion based joint
  video representation and ocelm.
\newblock {\em Neurocomputing}, 277:161--175, 2018.

\bibitem{wong2022aer}
L.~Wong, D.~Liu, L.~Berti-Equille, S.~Alnegheimish, and K.~Veeramachaneni.
\newblock Aer: Auto-encoder with regression for time series anomaly detection.
\newblock In {\em 2022 IEEE International Conference on Big Data (Big Data)},
  pages 1152--1161. IEEE, 2022.

\bibitem{xu2022anomaly}
J.~Xu, H.~Wu, J.~Wang, and M.~Long.
\newblock Anomaly transformer: Time series anomaly detection with association
  discrepancy.
\newblock In {\em International Conference on Learning Representations}, 2022.

\bibitem{9753580}
R.~K. Yadav and R.~Kumar.
\newblock A survey on video anomaly detection.
\newblock In {\em 2022 IEEE Delhi Section Conference (DELCON)}, pages 1--5.
  IEEE, 2022.

\bibitem{yan2018spatial}
S.~Yan, Y.~Xiong, and D.~Lin.
\newblock Spatial temporal graph convolutional networks for skeleton-based
  action recognition.
\newblock In {\em Proceedings of the AAAI conference on artificial
  intelligence}, volume~32, 2018.

\bibitem{yang2022recurring}
J.~Yang, X.~Dong, L.~Liu, C.~Zhang, J.~Shen, and D.~Yu.
\newblock Recurring the transformer for video action recognition.
\newblock In {\em Proceedings of the IEEE/CVF Conference on Computer Vision and
  Pattern Recognition}, pages 14063--14073, 2022.

\bibitem{yin2021graph}
W.~Yin, H.~Yin, D.~Kragic, and M.~Bj{\"o}rkman.
\newblock Graph-based normalizing flow for human motion generation and
  reconstruction.
\newblock In {\em 2021 30th IEEE International Conference on Robot \& Human
  Interactive Communication (RO-MAN)}, pages 641--648. IEEE, 2021.

\bibitem{yoo2018deep}
Y.~Yoo, L.~Y. Tang, T.~Brosch, D.~K. Li, S.~Kolind, I.~Vavasour, A.~Rauscher,
  A.~L. MacKay, A.~Traboulsee, and R.~C. Tam.
\newblock Deep learning of joint myelin and t1w mri features in
  normal-appearing brain tissue to distinguish between multiple sclerosis
  patients and healthy controls.
\newblock {\em NeuroImage: Clinical}, 17:169--178, 2018.

\bibitem{zhang2019view}
P.~Zhang, C.~Lan, J.~Xing, W.~Zeng, J.~Xue, and N.~Zheng.
\newblock View adaptive neural networks for high performance skeleton-based
  human action recognition.
\newblock {\em IEEE transactions on pattern analysis and machine intelligence},
  41(8):1963--1978, 2019.

\bibitem{zhou2023detecting}
Q.~Zhou, J.~Chen, H.~Liu, S.~He, and W.~Meng.
\newblock Detecting multivariate time series anomalies with zero known label.
\newblock In {\em Proceedings of the AAAI Conference on Artificial
  Intelligence}, volume~37, pages 4963--4971, 2023.

\bibitem{zou20203d}
S.~Zou, X.~Zuo, Y.~Qian, S.~Wang, C.~Xu, M.~Gong, and L.~Cheng.
\newblock 3d human shape reconstruction from a polarization image.
\newblock In {\em Computer Vision--ECCV 2020: 16th European Conference,
  Glasgow, UK, August 23--28, 2020, Proceedings, Part XIV 16}, pages 351--368.
  Springer, 2020.

\end{thebibliography}


\begin{comment}

\begin{IEEEbiography}{Michael Shell}
Biography text here.
\end{IEEEbiography}

\begin{IEEEbiographynophoto}{John Doe}
Biography text here.
\end{IEEEbiographynophoto}

%\newpage

\begin{IEEEbiographynophoto}{Jane Doe}
Biography text here.
\end{IEEEbiographynophoto}
\end{comment}

%\vfill


\end{document}